\documentclass[10pt,twocolumn,letterpaper]{article}

\usepackage{wacv}
\usepackage{times}
\usepackage{epsfig}
\usepackage{graphicx}
\usepackage{amsmath}
\usepackage{amssymb}
\usepackage{booktabs}
\usepackage{authblk}

\usepackage{algorithm}
\usepackage[noend]{algpseudocode}

\usepackage[accsupp]{axessibility}  

\def\wacvPaperID{65} 

\wacvapplicationstrack 

\wacvfinalcopy 

\ifwacvfinal
\usepackage[breaklinks=true,bookmarks=false]{hyperref}
\else
\usepackage[pagebackref=true,breaklinks=true,colorlinks,bookmarks=false]{hyperref}
\fi

\begin{document}

\title{ATCON: Attention Consistency for Vision Models}

\author[1,2]{Ali Mirzazadeh*}
\author[1]{Florian Dubost*}
\author[1]{Maxwell Pike\textsuperscript{1} \protect\\ Krish Maniar}
\author[2]{Max Zuo}
\author[1]{Christopher Lee-Messer}
\author[1]{Daniel Rubin}
\affil[1]{Stanford University}
\affil[2]{Georgia Institute of Technology \authorcr {\tt\small alimirz@gatech.edu, floriandubost1@gmail.com, \{cleemess, rubin\}@stanford.edu}}
\affil[*]{{\tt\small equal contribution}}

\maketitle
\thispagestyle{empty}

\begin{abstract}
   Attention--or attribution--maps methods are methods designed to highlight regions of the model’s input that were discriminative for its predictions. However, different attention maps methods can highlight different regions of the input, with sometimes contradictory explanations for a prediction. This effect is exacerbated when the training set is small. This indicates that either the model learned incorrect representations or that the attention maps methods did not accurately estimate the model’s representations. We propose an unsupervised fine-tuning method that optimizes the consistency of attention maps and show that it improves both classification performance and the quality of attention maps. We propose an implementation for two state-of-the-art attention computation methods, Grad-CAM and Guided Backpropagation, which relies on an input masking technique. We also show results on Grad-CAM and Integrated Gradients in an ablation study. We evaluate this method on our own dataset of event detection in continuous video recordings of hospital patients aggregated and curated for this work. As a sanity check, we also evaluate the proposed method on PASCAL VOC and SVHN. With the proposed method, with small training sets, we achieve a 6.6 points lift of F1 score over the baselines on our video dataset, a 2.9 point lift of F1 score on PASCAL, and a 1.8 points lift of mean Intersection over Union over Grad-CAM for weakly supervised detection on PASCAL. Those improved attention maps may help clinicians better understand vision model predictions and ease the deployment of machine learning systems into clinical care. We share part of the code for this article at the following repository: \href{https://github.com/alimirzazadeh/SemisupervisedAttention} {https://github.com/alimirzazadeh/SemisupervisedAttention}.
\end{abstract}

\section{Introduction}

\label{sec:intro}

In many real-world problems such as healthcare, labeled training data can be scarce \cite{park2019,chen2019b}, which can drive models to learn partially-incorrect representations and overfit to their training set \cite{xu2019,altman2018}. Consequently, in small datasets, researchers need to ensure that correct representations were learned. Those representations should match human understanding, be generalizable to unseen data, and not focus on potential bias in the dataset. This is especially essential to healthcare machine learning systems, where interpretability can justify predictions and decisions.

Attention--or attribution--map methods can be used the evaluate the representations of a model by highlighting regions in the input signal that are discriminative for the model's predictions. They have established themselves as one of the main methods for analyzing the interpretability of neural networks \cite{zhang2018}, and verify that the model did not leverage bias present in the data.
However, attention map results can greatly vary depending on the chosen attention computation method, to the point of being sometimes contradictory \cite{adebayo2018}. Moreover, some of these methods have been shown to be biased to irrelevant patterns in the data, such as regions of high-intensity gradients in images \cite{adebayo2018}. 
Attention maps also become more dissimilar when the task becomes more challenging and the chance of overfitting increases. For example, Dubost et al. For weakly supervised detection, \cite{dubost2020} show that there is a larger difference of performance between the Grad-CAM attention map \cite{selvaraju2017} and the Grad attention map \cite{simonyan2014} in datasets for which the overall performance is worse. 

Consequently, do models with harmonized attention, i.e. having similar attention maps computed from multiple methods, result in improved representations?

We propose to enforce consistency between attention maps computed using different methods to improve the representations learned by the model and increase its classification performance on unseen data. More specifically, we design an attention consistency loss function for two state-of-the-art attention maps methods: Grad-CAM \cite{selvaraju2017} and Guided Backpropagation \cite{springenberg2014}, but also show results for Grad-CAM and Integrated Gradients \cite{sundararajan2017} in the ablation study. We propose to optimize this loss function as an unsupervised fine-tuning step, to improve the representations of pretrained models.

We show that the proposed method (ATCON) can improve classification performance in video clip event classification with our own dataset curated for this project. The video dataset consists of clips extracted from continuous video recordings of hospital patients in their rooms. As a sanity check, we also show improvement in image classification with PASCAL VOC \cite{everingham10} (and SVHN \cite{netzer2011reading} in Appendix Section 2) when the size of the training set is reduced. We show that attention consistency improves the quality of attention maps. Attention maps are analyzed qualitatively on the video dataset, and quantitatively on PASCAL by computing the overlap between thresholded attention maps and ground truth bounding boxes. The benefits of the method are demonstrated for multiple network architectures: ResNet 50, Inception-v3, and a 3D 18 layers ResNet. We compare the method the baselines including layer attention consistency \cite{wang2019sharpen} and few-shot learning multi-label classification \cite{alfassy2019laso}. For the video dataset, we show that the proposed method can leverage the state-of-the-art self-supervised method SimCLR \cite{chen2020} to further boost performance. The improved attention maps may help clinicians better understand model predictions, ease the deployment of machine learning systems into clinical care, and eventually improve patients' outcomes.

\section{Related works}

A detailed description of related work on attention maps is given in Appendix Section 1. Below we detail related work in attention map consistency and semi/self-supervised learning.

\subsection{Attention Map Consistency}
To the best of our knowledge, consistency between multiple attention maps methods has not yet been used to evaluate and optimize a model's performance. The closest works entail inspecting the consistency of the same attention maps method under different augmented versions of the input \cite{guo2019}, or across the layers of the same network \cite{wang2019sharpen}. Li et al. \cite{li2020} extend the idea of Guo et al. \cite{guo2019} to attention consistency between image shared similar features. Xu et al. \cite{xu2020} combine the articles cited above to enforce consistency of attention maps both across augmented images and across network layers.

However, none of these articles inspect consistency across multiple attention map methods. The core mechanics of different attention map methods can be substantially different, and their ranking in terms of weakly supervised detection performance can greatly vary between datasets \cite{dubost2020}. In Appendix Section 1, we detail the major types of attention maps methods, including those utilized in this article, namely Grad-CAM \cite{selvaraju2017} and Guided Backpropagation \cite{springenberg2014}. 

\subsection{Semi-supervised and Self-supervised Learning}
When training labels are scarce, the go-to approach becomes unsupervised and semi-supervised learning, and especially self-supervised learning. 
In image classification, most state-of-the-art semi-supervised methods are based on self-supervision and use contrastive learning approaches \cite{he2020,chen2020}. 
Momentum Contrast (MoCo) \cite{he2020} encodes and matches query images to keys of a dynamic dictionary. SimCLR \cite{chen2020} improves upon MoCo by removing the need for specialized architectures. The authors of SimCLR stress that the composition of data augmentation is crucial in achieving high performance. Another key parameter is the batch size.
SimCLR works best with large batch sizes, which can become a strong limitation when working with high dimensional data such as video data. Jing et al. \cite{jing2021} proposed a semi-supervised learning method designed for video classification, using pseudo-labels and normalization probabilities of unlabeled videos to improve the classification performance.

Few-shot learning methods can also leverage small datasets by requiring only a few labeled samples. For example, FixMatch \cite{sohn2020} creates pseudo labels for unlabeled data using the current model's logits and include a cross-entropy term on those pseudo labels in the loss function. Prototypical network \cite{snell2017} is another few-shot learning method which allows networks to generalize to new classes with limited samples for those classes. Few-shot learning are rarely designed for multi-label classification problems, which prevent from using them on e.g. PASCAL. In this article, we compare our approach to LaSO \cite{alfassy2019laso}, a few-shot learning method for multi-label classification. 

Yet, most unsupervised, semi-supervised or self-supervised learning methods assume that a large quantity of (unlabeled) data are available, together with large computational resources to store and process them. In many real-world applications, such as medical applications, both data and computational resources are often lacking, especially for smaller institutions or emerging modalities. Acquisition costs can be high because of the price of the acquisition device, or because experimental settings deviate from clinical practice and require the setup of a dedicated research acquisition environment. For example, MRI scanners are expensive and only allow a limited number of patients to be scanned at once in an hospital. Experiments with MRI settings that target resolutions beyond 1mm isotropic are rarely used in clinical practice and require a specific environment. For example, Cicek et al. \cite{cciccek2016} published a prominent article in medical image analysis that only use 3 sparsely labeled 3D microscopy images in their study, and many medical image analysis competitions only provide a few scans for participants \cite{kuijf2019,timmins2021} also partially due to administrative cost relating to patient consent and personal health information.

In this article, we propose a regularization method that improves the classification and interpretability of neural networks on very small datasets, without requiring additional data.

\section{Method}

We have seen in the literature that attention maps become more dissimilar when the task becomes more challenging and the chance of overfitting increases \cite{dubost2020}. Consequently, we propose to improve the representations learned by classifiers by enforcing consistent representations across different types of attention maps. Firstly, we detail the concept of attention map consistency, and secondly, we propose an implementation for two state-of-the-art attention functions: Grad-CAM \cite{selvaraju2017} and Guided Backpropagation \cite{springenberg2014}. Note that the proposed method does not introduce any additional parameters.

\subsection{Attention Map Consistency}

Let us consider a set $X$ of $N$ samples $x_{n}$ with their corresponding labels $y_{n}$, and a classifier $f$ with parameters $\theta$, such that $f(x)=\hat{y}$. By definition, an attention function $g$ allows us to compute an attention map $A$ for a given input $x$ and classifier $f$ such that $g(f,x)=A_{x}$. The attention map $A$ has values in $\mathbb{R}$ and highlights the subset of $x$ that was most informative to the predictor, or that relates the most to its prediction. For example, if $x$ is an image, the attention map $A$ highlights areas of the images that are discriminative for the target prediction. While the dimensions of $A$ and $x$ do not have to be the same, we assume that there is at least a surjective mapping from $x$ to $A$, such that any element of the input $x$ can be linked to an element of $A$ (its discriminative power for the task).

Let us now consider a set of $M$ attention functions $g_{m}$ such that $g_{m}(f,x)=A_{x,m}$ for an input $x$. We want to maximize the correlation $h$ between the attention maps $A_{x,m}$ for the full set of input samples $x_{n}$ such that we seek to solve:

\begin{equation}
    \max_{\theta}\{\mathbb{E}_{x\in X} [h(A_{x,1},...,A_{x,M})]\}
\end{equation}

The correlation $h$ should be chosen to be high when the attention maps highlight similar regions of the input. The choice of the correlation function $h$ depends on the type of attention map to compare. We detail our choice of $h$ for Grad-CAM and Guided Backpropagation in section \ref{sec:gradcamAndGB} and propose ablation studies.

The corresponding attention map consistency loss function is defined as

\begin{equation}
L_{A} = -\sum_{x\in X} h(A_{x,1},...,A_{x,M}).
\end{equation}

\subsection{Training Strategy: Unsupervised Fine-Tuning}

The attention consistency loss does not require any training labels, which means that unsupervised training is possible. However, for the loss to converge to a desirable minimum, the classifier needs to have already learned meaningful representations prior to optimizing the consistency between attention maps. Hence, we propose the optimize the attention map consistency a posteriori, as an unsupervised fine-tuning step.

In the ablation study of the experiment section, we compare unsupervised fine-tuning to adding the consistency loss to the standard classification loss during training (categorical cross-entropy) by computing a linear combination of the two losses. We call this other strategy \textit{combined} optimization. We also compare to alternating between the two losses batch-wise as proposed by Chen et al. \cite{chen2019}. We call this last strategy \textit{alternated} optimization.

Further details about training hyperparameters, transformations, optimizers, and libraries are given in experiment section, Section \ref{sec:training}.

\subsection{Consistency between Grad-CAM and Guided Backpropagation}
\label{sec:gradcamAndGB}

We chose to implement the attention consistency loss using two state-of-the-art attention functions: Grad-CAM \cite{selvaraju2017} and Guided Backpropagation \cite{springenberg2014}. We selected these methods as they can be computed for any type of convolutional network and because their computation is different enough to witness significant changes in the attention consistency loss during training. 

Grad-CAM attention map $A_{Grad-CAM}$ is computed as a linear combination of the $k$ feature maps $f_{k}$ of a target convolutional layer:

\noindent\begin{minipage}{.5\linewidth}
\begin{equation}
\label{eq:grad-CAM}
A_{Grad-CAM} = \sum_{k}^{N}\alpha_{k}f_{k}
\end{equation}
\end{minipage}%
\begin{minipage}{.5\linewidth}
\begin{equation}
\alpha_{k} = \frac{1}{Z} \sum \frac{\partial \hat{y}}{\partial f_{k}},
\end{equation}
\end{minipage}

where each weight $\alpha_{k}$ is computed as the average of the derivative of the output $\hat{y}$ with respect to the feature maps $f_{k}$, and where $Z$ is the size of the feature map $f_{k}$. We choose the last convolutional layer of our network's architecture to compute Grad-CAM, as it is the most closely related to the output. Indeed, earlier layers often tends to be variants of object-agnostic edge detectors \cite{zeiler2014visualizing}.

Guided Backpropagation is computed by estimating the gradient of the network's output $\hat{y}$ with respect to the network's input $x$:

\begin{equation}
\label{eq:GB}
A_{GB} = \frac{\partial \hat{y}}{\partial x}.
\end{equation}

For multiclass classification problems, both Grad-CAM and Guided Backpropagation compute class-wise attention maps. We compute the attention map consistency loss only using the attention map of the top predicted class.

As mentioned in the related work section, Guided Backpropagation attention maps often have a higher resolution than that of Grad-CAM attention maps because of pooling layers in the architecture. Consequently, often there exists no bijective mapping between both attention maps which complicates the computations of the attention map correlation function $h$. Moreover, Grad-CAM and Guided Backpropagation tend to focus on semantically different regions of the input, such that a simple resizing operation would still incorrectly represent the correlation between the attention maps. 

To alleviate those issues, we propose to use a masking strategy. First, we compute Grad-CAM $A_{Grad-CAM,1}$ and Guided Backpropagation $A_{GB}$ attention maps. Then, a mask $P$ is derived from Guided Backpropagation to mask the input. The mask is computed according to the spatial attention masking defined by Wang et al. \cite{wang2017}: 

\begin{equation}
\label{eq:mask}
P(i) = \frac{1}{1 + \exp (-(A_{GB}(i)-\mu)/\sigma)},
\end{equation}

where $\mu$ is the mean over $A_{GB}$, $\sigma$ the variance, and $i$ spans the $A_{GB}$.
Subsequently, the forward propagation is run a second time using the masked input $P\odot x$. We compute Grad-CAM  $A_{Grad-CAM,2}$ a second time using the feature maps of this second forward propagation. 
Eventually, we compute the correlation between the two instances of Grad-CAM attention maps by vectorizing the attention maps and computing the Pearson correlation $L_{A}(\theta,x) = Pearson(A_{Grad-CAM,1},A_{Grad-CAM,2})$.  Correlating the two Grad-CAM maps $A_{Grad-CAM,1}$ and $A_{Grad-CAM,2}$ indirectly correlates the original Grad-CAM map $A_{Grad-CAM,1}$ to the Guided Backpropagation map $A_{GB}$ because the masking with $A_{GB}$ forces $A_{Grad-CAM,2}$ to highlight regions already highlighted by $A_{GB}$ itself.

This masking method was inspired by perturbation methods for attention map computation \cite{petsiuk2018}, and can be generalized to compute attention map consistency with any type of attention map. Algorithm \ref{alg:cap} summarizes the masking and computation of the attention consistency loss function.

\begin{algorithm}
\caption{Proposed attention consistency for Grad-CAM and Guided Backpropagation}\label{alg:cap}
\hspace*{\algorithmicindent} \textbf{Input:} sample $x$, convolutional neural network $f$ \\
\hspace*{\algorithmicindent} \textbf{Output:} attention consistency loss $L_{A}$

\begin{algorithmic}[1]
\State \textit{Forward propagate $x$ through $f$}
\State $A_{Grad-CAM, 1} \gets \textit{Grad-CAM}$ (Equation \ref{eq:grad-CAM}) 
\State $A_{GB} \gets \textit{Guided Backpropagation}$ (Equation \ref{eq:GB}) 
\State $P \gets \textit{mask computed from } A_{GB}$ (Equation \ref{eq:mask}) 
\State $x_{masked} \gets P\odot x$
\State \textit{Forward propagate $x_{masked}$ through $f$}
\State $A_{Grad-CAM, 2} \gets \textit{Grad-CAM computed from } x_{masked}$ (Equation \ref{eq:grad-CAM}) 
\State $L_{A} \gets Pearson(A_{Grad-CAM,1},A_{Grad-CAM,2})$

\end{algorithmic}
\end{algorithm}

\section{Experiments}

We show that our method, forcing attention consistency alone a posteriori, can improve classification performance, while improving the representations of the classifier. We show that the performance gain varies with the training set size, and we show the benefits of the method for three architecture: ResNet 50 \cite{he2016}, Inception-v3 \cite{szegedy2016}, and 3D 18 layers ResNet \cite{tran2018}. The proposed method is evaluated on a real-world dataset, which we curated for this project: event recognition in continuous recordings of hospital patients. Sanity checks are performed on the PASCAL-VOC dataset \cite{everingham10}. Additional experiments on SVHN \cite{netzer2011reading} are presented in Appendix Section 2. We compare the method the baselines including layer attention consistency \cite{wang2019sharpen} and few-shot learning multi-label classification \cite{alfassy2019laso}. 
We also present an three ablation studies. First, attention consistency with Grad-CAM and Guided Backpropagation is computed using four different resolution matching strategies and three different correlation measures. Second, we compare unsupervised fine-tuning, to combined and alternated optimization. Third, we compared the optimizing consistency between Grad-CAM and Gruided-backpropagation to consistency between Grad-CAM and Integrated Gradients.

\subsection{Datasets}
We aggregated and curated a dataset of continuous video recordings of hospital patients in an epilepsy center unit for children and neonates, collected with IRB oversight and approval. We identified video clips displaying five types of events: patting of the neonates by nurses, suctioning of neonates' mouth liquid by nurses, rocking of neonates, patient chewing food, and finally cares being done on the patient by nurses. Those events are selected as they can mislead automated seizure detection systems. Clips are post-processed to be 4 sec long, sampled at 4 frames per second. Consequently, the task is defined as a five-classes video clip classification problem. The curated dataset includes 59 patients and 2 hours 18 minutes of video recordings, with a median event clip length of 25 seconds. The frame resolution is reduced from 320x240 to 80x80 using subsampling. The video clips were separated into three balanced sets of similar size, using data of different patients for each sets. The first set has 32 4-sec clips per class, the second set 49 and the last set 35. More statistics about the dataset can be found in Appendix Section 3.

PASCAL VOC \cite{everingham10} is used for sanity check and is reframed as a 20 classes multi-label classification dataset. The bounding box ground truth annotations are not used for training. They are only used during inference to evaluate the weakly supervised detection capability of the generated attention maps. 500 random images sampled from the 5717 images of the training set are used a validation set, and a varying number of images, from 2 to 16 images per class, are sampled from the remaining set and used for training. The publicly released validation set of 5823 images is left out during training, and used a separate test set to evaluate the methods.

\begin{table*}[t]
\caption{\textbf{Classification results on the video dataset.} \textit{We compare the baseline method, the baseline fine-tuned with the proposed attention consistency (B + ATCON), SimCLR pretraining  \cite{chen2020} (SimCLR), and SimCLR pretraining fine-tuned with the proposed attention consistency (S + ATCON). The first four rows indicate models trained with 16 samples per class, and the last four rows with 32 samples per class. We show F1 scores rescaled to [0,100]. Mean F1 is the F1 averaged over the five classes. F1 suctioning, chewing, rocking, cares and patting show class-wise F1s. Bootstrapped confidence intervals are indicated in brackets. The highest performance is indicated in bold.} \\\hspace{\textwidth}
}
\resizebox{\textwidth}{!}{%
\begin{tabular}{c|c|ccccc}
Method        & Mean F1              & F1 suctioning        & F1 chewing           & F1 rocking           & F1 cares             & F1 patting           \\ \hline
Baseline -- 16 samples    & 15.5 {[}11.6-19.4{]} & 37.9 {[}25.0-49.3{]} & 0.0 {[}0.0-0.0{]}    & 0.0 {[}0.0-0.0{]}    & \textbf{30.7 {[}21.7-39.7{]}} & \textbf{9.3 {[}0.0-20.0{]}}   \\
B + ATCON & 17.4 {[}13.3-21.9{]} & 42.1 {[}31.2-52.2{]} & 8.2 {[}0.0-16.2{]}   & 16.4 {[}4.8-29.0{]}  & 20.2 {[}11.6-28.6{]} & 0.0 {[}0.0-0.0{]}    \\
SimCLR        & 23.1 {[}18.8-27.7{]} & 47.1 {[}35.1-58.3{]} & 27.6 {[}16.3-38.9{]} & 17.1 {[}6.6-29.2{]}  & 23.8 {[}14.7-33.3{]} & 0.0 {[}0.0-0.0{]}    \\
S + ATCON & \textbf{29.7 {[}25.2-34.3{]}} & \textbf{50.4 {[}37.5-62.2{]}} & \textbf{40.7 {[}29.4-51.4{]}} & \textbf{43.2 {[}32.5-53.9{]}} & 14.5 {[}5.5-24.3{]}  & 0.0 {[}0.0-0.0{]}    \\ \hline
Baseline -- 32 samples      & 26.5 {[}21.6-31.7{]} & 30.1 {[}19.7-40.4{]} & 36.4 {[}24.7-46.9{]} & 9.3 {[}0.0-20.5{]}   & 44.0 {[}33.0-54.5{]} & 13.0 {[}0.0-25.0{]}  \\
B + ATCON & 27.5 {[}22.7-32.8{]} & \textbf{33.6 {[}23.5-43.0{]}} & \textbf{46.7 {[}35.9-57.1{]}} & 5.4 {[}0.0-15.8{]}   & 29.2 {[}16.7-41.0{]} & \textbf{23.2 {[}10.9-35.7{]}} \\
SimCLR        & 29.3 {[}23.8-34.8{]} & 21.3 {[}10.5-32.0{]} & 43.9 {[}30.2-56.5{]} & \textbf{30.9 {[}18.2-43.5{]}} & 42.0 {[}31.2-51.9{]} & 7.7 {[}0.0-16.7{]}   \\
S + ATCON & \textbf{31.4 {[}25.9-37.0{]}} & 31.2 {[}17.2-43.3{]} & 35.4 {[}22.9-47.5{]} & 30.5 {[}17.5-42.9{]} & \textbf{48.0 {[}38.3-57.6{]}} & 12.2 {[}3.6-23.3{]} 
\end{tabular}
}
\label{table:video_f1}
\end{table*}

\begin{table*}[]
\centering
\caption{
\textbf{Classification Results on PASCAL.} \textit{We compare the classification results of ResNet without data augmentation (ResNet no Aug ), ResNet with data augmentation (ResNet),  ResNet with data augmentation (ResNet) and layer attention consistency (ResNet  + Layer Att), and ResNet with data augmentation and the proposed attention consistency fine-tuning (ResNet + ATCON). Results are shown for models trained with varying numbers of training samples per class. Stars indicate significant differences.} \\\hspace{\textwidth}
}
\resizebox{15cm}{!}{%
\begin{tabular}{c|cccccc|cccccc}
Methods              & \multicolumn{6}{c|}{F1, Training Sample per Class}                                          & \multicolumn{6}{c}{mAP, Training Sample per Class}                                            \\ \cline{2-13} 
                     & 2             & 4             & 8             & 12            & 16          & 135           & 2             & 4             & 8             & 12            & 16            & 135           \\ \hline
ResNet no Aug   \cite{he2016}           & 23.9          & 43.1          & 57.4          & 59.9          & 64.7        & 76.9          & 51.4          & 65            & 73.2          & 76.9          & 78.6          & 85.2          \\
ResNet  \cite{he2016}     & 38.3          & 46.9          & 60.2          & 64.1          & 68.3 & \textbf{77.6} & 57.4          & 67.2          & 74.4          & 77.5 & 79.4 & \textbf{85.7} \\
ResNet  + Layer Att \cite{wang2019sharpen}   & 37.8         & 48.8          & 60.4          & 64.1          & \textbf{68.6} & \textbf{77.6} & 57.2          & 67.0          & 74.4          & \textbf{77.6} & \textbf{79.5} & \textbf{85.7} \\
ResNet + ATCON & \textbf{41.2*} & \textbf{51.6*} & \textbf{62.5*} & \textbf{64.8} & 68.2        & 77.3          & \textbf{58.1} & \textbf{67.9} & \textbf{74.9} & 77.2          & \textbf{79.4}          & \textbf{85.7}
\end{tabular}
}
\label{table:pascal_ap}
\end{table*}

\subsection{Models and Training}
\label{sec:training}
For the experiements on hospital videos, we use a 3D 18 layers ResNet \cite{tran2018} pretrained on Kinetics-400 \cite{kay2017}. For the experiments on PASCAL, we use state-of-the-art 2D networks: ResNet 50 \cite{he2016} and Inception-v3 \cite{szegedy2016} (only for PASCAL), both pretrained on ImageNet \cite{deng2009}. 

On top of ImageNet or Kinetics pretraining, we train on PASCAL or the video dataset using $N$ samples per class and a categorical cross-entropy loss function and the Adam optimizer with a learning rate of $0.001$. Then, we perform unsupervised fine-tuning with the proposed method using the same training samples. The batch size is 4 for PASCAL, and 12 for the video dataset. We select the model that maximizes the mean average precision on PASCAL (because it is multi-label) and the mean F1 score on the video dataset, computed over the validation set. The test is completely left out of the training loop for final evaluation.

The models are regularized with data augmentation. 
For the video dataset, data augmentation includes random color jitters with brightness, contrast, and saturation up to 0.8, and hue up to 0.4, random crops of 90 percent of the image size in x and y, and random horizontal flips.
For PASCAL, data augmentation includes random rotations up to 10 degrees, 50 percent chance of horizontal or vertical flip, and resizing to 256x256 for ResNet or to 299x299 for Inception-v3 due to architecture constraints.

Our code is available online on GitHub at \textit{removed for blind review}, and builds on PyTorch 1.9.0 and Torchvision 0.10.0. We used two NVIDIA Titan RTX for training.

\subsection{Classification Results}
On the video dataset, we compare the proposed method, ATCON, to training a 3D 18 layers ResNet without attention consistency.
We also include a comparison to SimCLR pretraining \cite{chen2020} using the same training samples, followed by the same standard training, without attention consistency. All training procedures use the same training samples. We vary the number of training samples per class from 16 to 32. Table \ref{table:video_f1} shows the F1 scores. The proposed attention consistency method provides a lift of mean F1. This lift increases when SimCLR is used for pretraining.

To verify that these results generalize to other datasets, we repeat these experiments on PASCAL. We compare ATCON to training ResNet 50 without attention consistency. We also compare to not using data augmentation during this training stage. Lastly, we compare ATCON to the layer attention consistency method proposed by Wang et al. \cite{wang2019sharpen}.
As recommended by the authors, we use the last convolution layers of the last two blocks in ResNet to compute the attention consistency. The layer attention consistency is implemented as a fine-tuning step, which shows better results than simultaneous training with categorical cross-entropy.
We repeat the experiments with varying numbers of training samples from 2 to 135 (the maximum possible) per class (2, 4, 8, 12, 16 and 135). The F1 scores and mean average precision on the test are shown in Table \ref{table:pascal_ap}. Experiments on PASCAL confirm the findings of the curated video dataset, and show that fine-tuning can improve classification performance when training data is scarce. ATCON outperforms the layer attention consistency baseline \cite{wang2019sharpen}, and combining both methods did not show higher results than the proposed attention consistency method alone.

\begin{figure*}[t!]
\centering
\includegraphics[width=14cm]{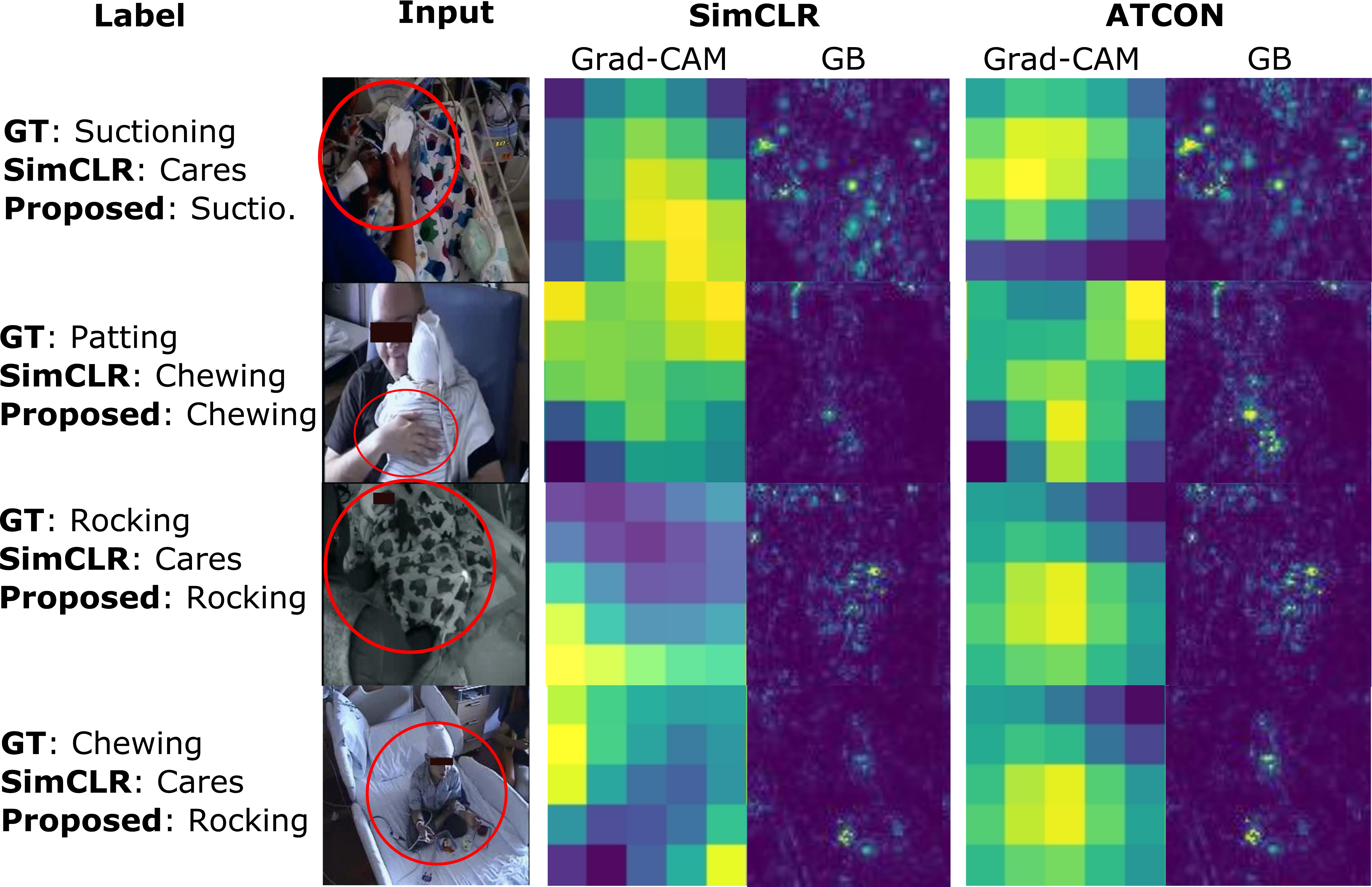}
\caption{
\textbf{Examples of attention maps for hospital video data.} Comparison between SimCLR and SimCLR plus the proposed unsupervised attention consistency fine-tuning (ATCON). The first column indicates the ground truth (GT) label, the label predicted by SimCLR, and by the proposed method. The second column shows one of 16 input frames. We circled in red the region where the event is occurring. The two middle columns show Grad-CAM and Guided Backpropagation (GB) for SimCLR, and the last two columns show the same attention maps for the proposed method. Models were training with 16 samples per class. Examples showing substantial differences between methods were selected.
}
\label{fig:video_exmaples}
\end{figure*}

\begin{figure*}[t!]
\centering
\includegraphics[width=13cm]{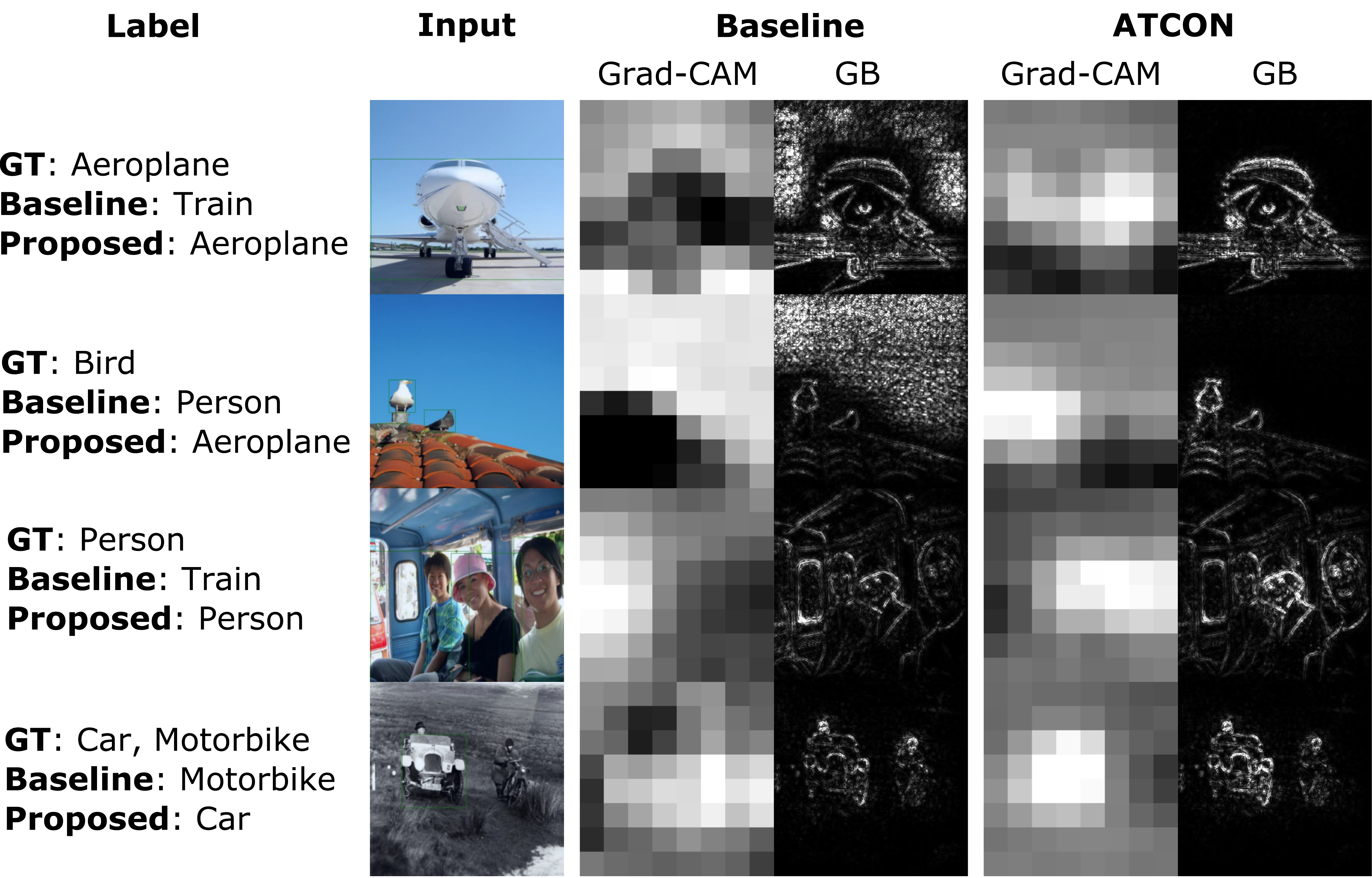}
\caption{
\textbf{Examples of attention maps for PASCAL.} Comparison between the baseline ResNet with data augmentation (Baseline) and the same method plus the proposed unsupervised attention consistency fine tuning (ATCON). The first column indicates the ground truth (GT) label, the label predicted by the baseline, and by the proposed method. The second column shows the input frames. The two middle columns show Grad-CAM and Guided Backpropagation (GB) for the baseline, and the last two columns show the same attention maps for the proposed method. Models were training with 4 samples per class. Examples with clear differences between methods were selected for display.
}
\label{fig:pascal_examples}
\end{figure*}

\begin{table}[t]
\centering

\caption{
\textbf{Overlap on PASCAL.} \textit{Intersection over Union between Grad-CAM attention maps and ground truth bounding boxes. The overlap is computed only for true positives classifications. Results are shown for Inception with data augmentation (Inception), Inception with data augmentation and the proposed attention consistency fine-tuning (Inception + ATCON), ResNet without data augmentation (ResNet no Aug), ResNet with data augmentation (ResNet), ResNet without data augmentation (ResNet no Aug), ResNet with data augmentation and layer attention consistency (ResNet + Layer Att), and ResNet with data augmentation and the proposed attention consistency fine-tuning (ResNet + ATCON). Each column corresponds to a model trained with a varying number of samples per class. The highest performance is indicated in bold. Stars indicate significant statistical improvement.} \\\hspace{\textwidth}
}

\resizebox{8cm}{!}{
\begin{tabular}{c|cccccc}
Method                & \multicolumn{6}{c}{Training Samples per Class}                                                \\ \cline{2-7} 
                      & 2             & 4             & 8             & 12            & 16            & 135           \\ \hline
Inception    \cite{szegedy2016}   &               & 44.5          & 44.1          & 44.5          & 43.9          & \textbf{43.3} \\
Inc + ATCON &               & \textbf{44.6} & \textbf{44.2} & \textbf{44.9} & \textbf{44.6} & \textbf{43.3} \\ \hline
ResNet no Aug  \cite{he2016}               & 42.8          & 46.6          & 47.7          & 47.6          & 48.1          & \textbf{45.2}          \\
ResNet    \cite{he2016}       & 44.6            & 50.0          & 48.5          & 49.0          & 47.9          & 44.9          \\
ResNet + Layer Att  \cite{wang2019sharpen}        & 43.2           & 47.8          & 47.1          & 47.3          & 46.9          & 43.6          \\
ResNet + ATCON   & \textbf{46.4*} & \textbf{50.2*} & \textbf{49.5} & \textbf{49.3} & \textbf{48.4} & 45.1
\end{tabular}
}
\label{table:overlap}

\end{table}

\subsection{Learned Representations}
We consider the generated attention maps as a measure of the correctness of the representations. If the attention maps highlight the target objects in the image, we expect the model to have learned correct representations. We visualize attention maps of both the video dataset and PASCAL in Figures \ref{fig:video_exmaples} and \ref{fig:pascal_examples}. We also plot the gain of attention consistency with respect to the original attention consistency on PASCAL (Appendix Section 1).

There were no bounding box annotations to quantify the improvement of the attention maps localization on the video dataset. Instead, we perform the analysis on PASCAL. Using the intersection of union, we compute the overlap of the ground truth bounding boxes with Grad-CAM attention maps rescaled in $[0,1]$ and thresholded at 0.5. We compute the overlap only for true positive image classifications (the number of true positive images increases with the classification performance and the number of training samples per class).
Table \ref{table:overlap} shows the overlap for varying number of training samples for ResNet, Inception-v3, and for the layer attention consistency baseline \cite{wang2019sharpen}. The overlap is always higher with the proposed method, for all training sample sizes and both network architectures. We also notice that forcing layer attention consistency \cite{wang2019sharpen} reduces the localization capacity of the model. Contrary to the proposed method, with layer attention consistency \cite{wang2019sharpen}, the performance gain observed in Table \ref{table:pascal_ap} for smaller datasets is probably due to the induced smaller variance of the network weights than more accurate attention maps.

Qualitative inspection of the attention maps on PASCAL (Figure \ref{fig:pascal_examples}) and the video dataset (Figure \ref{fig:video_exmaples}) also reveal that attention consistency can improve the attention maps to focus on the target subject, while still predicting the incorrect label. Such an improvement has not been accounted for in our quantitative measure of the overlap (Table \ref{table:overlap}) as only true positives were utilized.

Note that attention consistency displayed smaller improvements on Inception-v3 than on ResNet (Table \ref{table:overlap}). This may indicate that Inception-v3 directly learns better representations. Inception was designed to be computationally efficient, with a reduced number of parameters. This may explain why the network seems to learn more generalizable representations \cite{szegedy2016}.

\subsection{Comparison to SOTA few-shot learning}

We also compare our method to LaSO \cite{alfassy2019laso}, a few-shot learning multi-label classification method. Experiments are performed on PASCAL and the mean-average precision is computed on the validation set (Table \ref{table:sota2}, details in Appendix Section 5). 

\begin{table}[]
\centering
\caption{
\textbf{Few-shot learning.} \textit{Comparison with the few-shot learning method LaSO\cite{alfassy2019laso} on PASCAL. mAP on the validation set.} \\\hspace{\textwidth}
}
\resizebox{6cm}{!}{%
\begin{tabular}{c|c| c |c|c|c}
Imgs Per Class   & 2                             & 4     & 8     & 12    & 16    \\ \hline
LaSO \cite{alfassy2019laso}        & 60.3 & 68.9 & 73.2 & - & - \\
ATCON        & 65.5 & 73.3 & 78.1 & - & - \\ 
\end{tabular}
}
\label{table:sota2}
\end{table}

\section{Ablation Study}

\subsection{Attention Consistency loss function}
To justify our choice of attention consistency loss function, we present an ablation study varying its two main components: the resolution matching technique and the correlation metric used to estimate attention map overlap. We consider that an attention consistency loss function is good if a lower consistency loss is equivalent to a lower supervised classification loss--the cross-entropy in our experiments. To quantify this relationship, we measure the correlation between both the unsupervised consistency loss and the supervised classification loss during a fully supervised training procedure. In this experiment, only the fully supervised classification loss is used to update the network weights. The unsupervised consistency loss is only monitored.

We explore three correlation measures: Structural Similarity Index Measure (SSIM), cross-correlation and Pearson coefficient; and four resolution matching techniques: the proposed masking technique using Guided Backpropagation to create the mask, the proposed masking technique using Grad-CAM to create the mask, smoothing and upsampling of Grad-CAM with linear interpolation, and downsampling of Guided Backpropagation with max pooling.

Table \ref{table:ablation} shows the results of the ablation study on PASCAL. The best combination of resolution matching strategy and correlation measure was Pearson and Guided Backpropagation as a mask. The second best was SSIM and Guided Backpropagation as a mask. Using Grad-CAM upsampling or Guided Backpropagation pooling for resolution matching was worse than Guided Backpropagation masking but still gave satisfying results, independently of the correlation metrics. Using Grad-CAM as a mask for resolution matching should be avoided.
 
 \subsection{Training Strategies}
We also compare the three different training strategies: optimizing the consistency loss as an unsupervised fine-tuning step (fine-tuning), together as a linear combination with the supervised loss (combined), and alternating between both losses batch-wise during training (alternated). Experiments are done on PASCAL with ResNet 50 and 4 training samples per class. The combined approach gets an attention map overlap of 49.3\% IoU and a F1 of 48.9\%. The alternated one 49.5\% IoU and a F1 of 50.6\%. As reported earlier, the fine-tuning approach got an 50.2\% IoU and a F1 of 51.6\%.

\subsection{Types of Attention Maps}
In the rest of the article, we report results using the proposed attention consistency between Grad-CAM and Guided-backpropagation. The proposed method reaches similar performance using Grad-CAM and Integrated Gradients \cite{sundararajan2017}, another attention method (Table \ref{table:sota1}).

\begin{table}[t]
\centering
\caption{
\textbf{Ablation study showing the Pearson correlation between different attention consistency losses and the target supervised loss.} \textit{High correlations show that the attention consistency loss estimates the target supervised loss well. The training was realized with a supervised ResNet for 100 epochs on PASCAL, and the supervised cross-entropy loss was computed on the validation set. 135 samples per class, the maximum in our classification dataset, was used for this experiment. The rows indicate the resolution matching techniques, and the columns the correlation measures.} 
}
\resizebox{8cm}{!}{
\begin{tabular}{c|ccc}
   & Pearson    & Cross-correlation    & SSIM    \\ \hline
Grad-CAM Upsampling & 65.4  & 67.5  & 60.3  \\
GB Pooling & 65.5  & 64.1  & 68.8  \\
GB as mask & 82.6  & 26.7  & 80.6  \\
Grad-CAM as mask & -51.9 & -67.2 & -75.8
\end{tabular}
}
\label{table:ablation}
\end{table}

\begin{table}[]
\centering
\caption{
\textbf{Types of Attention Maps.} \textit{Classification (mF1 and mAP) and overlap results for Grad-CAM with Integrated Gradients \cite{sundararajan2017} (instead of Grad-CAM with Guided Backpropagation) on PASCAL. \textit{m} is the number of  integrated gradients steps.} \\\hspace{\textwidth}
}
\resizebox{6cm}{!}{%
\begin{tabular}{c|ccccc}
Imgs Per Class   & 2                             & 4     & 8     & 12    & 16    \\ \hline
\textbf{mF1}     &                               &       &       &       &       \\
 m = 5         & 40.0                           & 47.5 & 60.1 & 64.4 & 68.5 \\
 m = 10        & 38.9                         & 46.8 & 60.3 & 64.8 & 68.4 \\ \hline
\textbf{mAP}     &                               &       &       &       &       \\
 m = 5         & 57.2                         & 67.2 & 74.3 & 77.5 & 79.5 \\
 m = 10        & 57.4 & 67.1 & 74.5 & 77.6 & 79.3 \\ \hline
\textbf{Overlap} &                               &       &       &       &       \\
 m = 5         & 44.8                         & 47.5 & 47.2 & 47.4 & 47.0  \\
 m = 10        & 44.0                          & 47.6 & 47.1 & 47.5 & 46.8
\end{tabular}
}
\label{table:sota1}
\end{table}

\section{Limitations}

The proposed method is only beneficial with small training sets, with larger training sets there is no significant improvement: neither in classification nor detection performance (see Appendix Section 6) for additional discussion).

Improving attention consistency was challenging for samples whose original attention maps did not overlap. Introducing a stochastic process in the computation of the attention consistency may allow attention maps to overlap and help the network converge to the target solution.

We refer to our training strategy as unsupervised fine-tuning, but we still monitor the generalization performance using the labels of the validation set. Knowing when to stop training without utilizing any labels could have substantial practical implications, where networks can be fine-tuned for attention consistency on deployment dataset without labels.

We evaluated the layer attention consistency baseline \cite{wang2019sharpen} in PASCAL and did not observe quantitative improvements in the classification performance or in attention maps quality, even combined with ATCON. While we may likely observe similar results in the video dataset, we have not performed the analysis.

\section{Conclusions}
We proposed ATCON, a method to optimize the consistency of attention maps, and propose an implementation for Grad-CAM and Guided Backpropagation. We showed on our own video dataset and on PASCAL that the method can improve both classification performance and the quality of attention maps. The proposed method contributes to research in network interpretability and application in small datasets.

{\small
\bibliographystyle{ieee_fullname}
\bibliography{egbib}
}

\end{document}